\documentclass[sigconf]{acmart}

\usepackage{booktabs}
\usepackage{array}
\usepackage{pgfplots}
\pgfplotsset{compat=1.18}
\AtBeginDocument{%
  }

\setcopyright{none}
\renewcommand\footnotetextcopyrightpermission[1]{}

\copyrightyear{2026}
\acmYear{2026}
\acmDOI{}
\acmConference[ACM AI Summit '26]{ACM AI Leadership Summit 2026}{August 30--September 2, 2026}{Atlanta, GA, USA}
\acmISBN{}

\begin{document}

\title[Confident but Unreliable: Auditing VLMs on Brain MRI]{Confident but
Unreliable: A Behavioral Safety Audit of Vision--Language Models on Brain MRI}








\author[Amir Sabbaghziarani et al.]{%
\texorpdfstring{%
  \begin{tabular}{@{}c@{\hspace{2.5em}}c@{\hspace{2.5em}}c@{}}
    \LARGE {Amir Sabbaghziarani} &
    \LARGE {Mohammadsajad Abavisani} &
    \LARGE {Sergey Plis}
    \\
    \large {TReNDS Center} &
    \large {TReNDS Center} &
    \large {TReNDS Center}
    \\
    \large {Atlanta, GA, USA} &
    \large {Atlanta, GA, USA} &
    \large {Atlanta, GA, USA}
  \end{tabular}
  \\[10pt]
  \parbox{0.75\textwidth}{%
    \centering\large
Tri-Institutional Georgia State University/Georgia Institute of Technology/Emory University Center for Translational Research in Neuroimaging and Data Science (TReNDS)
  }
}{Amir Sabbaghziarani; Mohammadsajad Abavisani; Sergey Plis}%
}

\renewcommand{\shortauthors}{Sabbaghziarani et al.}

\begin{abstract}
Vision--language models (VLMs), including medical specialists, are increasingly
proposed for medical imaging, yet their stated confidence is rarely evaluated
separately from correctness. We use brain MRI as a controlled, high-stakes testbed
for a broader failure mode in frontier multimodal systems: models can appear
competent while lacking reliable self-knowledge. We present an automatically
graded behavioral audit and pilot study of six instruction-tuned VLMs (five
general-purpose and one medical specialist) on 4{,}102 images---4{,}032
axial/coronal/sagittal MRI slices from 250 subjects plus 70 non-brain/noise
controls---with labels derived from public metadata and released expert
segmentation masks rather than new human annotation.

Across models, answer coverage is near-complete, but verbalized-confidence
calibration is poor: ECE ranges from 0.27 to 0.40, mean confidence on incorrect
answers ranges from 0.82 to 0.97, and 33--46\% of answered items are
high-confidence errors. The most accurate model is also the most confident on its
errors, while a base/specialist family contrast suggests that medical adaptation
improves tumor-presence detection without improving confidence reliability.
Open-ended diagnostics further show that hallucination and abstention vary
separately from multiple-choice accuracy. These findings argue that medical-image
VLM evaluation should report verbalized-confidence reliability, confident error,
hallucination, and abstention alongside accuracy. The code is publicly available at
\url{https://github.com/amir-sbg/confident-unreliable-vlm-brain-mri-audit}.
\end{abstract}



\maketitle

\section{Introduction}
General-purpose and medical vision--language models (VLMs) produce fluent,
clinically-styled descriptions of medical images and are increasingly proposed as
triage aids, ``second readers,'' or educational tools. Fluency, however, is not
competence. A model that answers confidently but incorrectly---and that never
declines to answer---is a safety hazard precisely \emph{because} it sounds
authoritative: a clinician or downstream system has no signal that the output
should be distrusted. Standard benchmarks emphasize accuracy and seldom ask the
questions that matter for safe deployment: \emph{does a model's stated confidence
track whether it is right? does it abstain on inputs it cannot interpret? when it
is wrong, how confident is it?}

Brain MRI is an attractive substrate for studying these questions at scale because
much of its ground truth can be derived automatically and without human
annotation: the imaging \emph{sequence} (T1, T1ce, T2, FLAIR, PD) is recorded in
acquisition metadata; the imaging \emph{plane} is fixed by how a slice is
extracted from the volume; whether an image \emph{is a brain MRI} follows from the
data source (and can be probed with deliberately out-of-distribution controls);
and \emph{tumor presence} and \emph{laterality} follow from expert segmentation
masks evaluated on the exact extracted slice. This lets us grade tens of thousands
of (image, question) pairs deterministically and foreground the safety-relevant
quantities---calibration and confident error---rather than accuracy alone. We
emphasize that this is a behavioral safety audit on public, brain-MRI-derived
slices; we make no claim about clinical diagnostic performance. Although the
substrate is brain MRI, the underlying question is general: whether a frontier
multimodal model's stated confidence tracks its competence in a high-stakes
visual domain where errors are costly.

\noindent\textbf{Contributions.}
\begin{itemize}
\item A reproducible audit protocol requiring \emph{no new annotation} for VLM reliability on
brain MRI, with slice-accurate labels from metadata and segmentation masks and a
metric suite---coverage, balanced accuracy, ECE/Brier, mean confidence on wrong
answers, confidently-wrong rate, and abstention---paired with subject-clustered uncertainty estimates for robustness.
\item Pilot evidence across six models that \emph{confidence is decoupled from
competence}: coverage is near-complete and calibration is poor, and---critically
---the most accurate model audited is the \emph{most} overconfident on its
errors, so improvements in raw accuracy did not yield improvements in
reliability.
\item A \emph{family-level base-vs-medical} comparison: medical adaptation was
associated with improved tumor-presence detection but reduced performance on some
generic visual tasks, with no improvement in verbalized-confidence reliability.
\end{itemize}

\section{Related Work}
Modern neural networks are known to be miscalibrated, typically
overconfident~\cite{guo2017calibration}. In the medical domain, recent work shows
that overconfidence persists across VLM families and scales and is not removed by
prompting strategies such as chain-of-thought or verbalized
confidence~\cite{byun2026overconfidence}, and that a model's confidence frequently
fails to separate correct from incorrect answers and is largely robust to
question rephrasing~\cite{khanmohammadi2026calibrated}. Large VLMs are also prone
to asserting content not supported by the image~\cite{li2023pope}. Separately,
medical multimodal LLMs can \emph{underperform} their general base models on image
tasks, with the degradation traceable to the visual-representation
pipeline~\cite{zhu2026lost}. We build on these observations by (i) grading fully
automatically and slice-accurately on brain MRI drawn from the BraTS
lineage~\cite{deverdier2024brats,menze2015brats,bakas2017advancing} and the IXI
healthy-volunteer cohort~\cite{ixi}; (ii) testing whether the pattern persists in
the newest, strongest model; and (iii) probing the effect of medical
adaptation through a base/specialist family contrast. We quantify calibration
with the Expected Calibration Error~\cite{naeini2015obtaining} and the Brier
score~\cite{brier1950verification}, report balanced
accuracy~\cite{brodersen2010balanced} because several tasks are class-imbalanced,
and compute subject-clustered intervals with the
bootstrap~\cite{efron1993bootstrap}.

\section{Audit Design}
\textbf{Data.} We assemble an exam set of 4{,}102 images---4{,}032 MRI slices from 250 subjects plus 70 non-brain/noise controls---across
three sources. BraTS-2024 post-treatment glioma scans (1{,}350 slices, 150
subjects) supply tumor-bearing volumes with expert segmentation
masks~\cite{deverdier2024brats}; the IXI dataset (2{,}682 slices, 100 subjects)
supplies healthy-volunteer scans with no tumor~\cite{ixi}; and 70 negative-control
images (non-brain photographs and synthetic noise) probe whether a model abstains
on inputs that are not interpretable brain MRI. Slices are extracted at controlled
fractional depths (0.4, 0.5, 0.6) within each volume, intensity-normalized, and
resized to a common size.

\noindent\textbf{Labels without new annotation.} Labels are derived without new human annotation: \emph{sequence} from acquisition metadata; \emph{plane} from the
extraction geometry; \emph{is-a-brain-MRI} from the data source; and \emph{tumor
presence and laterality} from the segmentation mask \emph{of the exact extracted
slice}. A slice is labeled tumor-positive only if its in-slice mask contains at
least 25 lesion voxels; laterality is computed from the in-slice lesion centroid
relative to the mid-sagittal line. This slice-level treatment is essential:
because most slices of a tumor-bearing subject contain no visible tumor,
subject-level labels would penalize a model that correctly reports ``no lesion''
on a lesion-free slice. Under the visibility threshold the slice-level label
distribution is 3{,}118 tumor-negative and 840 tumor-positive slices; 144 slices
without an available mask are excluded from tumor/laterality grading.

\noindent\textbf{Tasks.} Each interpretable slice is queried on five
auto-gradable tasks (chance shown in parentheses): T1 sequence identification
(7-way, $0.143$); T2 plane (axial/coronal/sagittal, $0.333$); T3 is-this-a-brain
($0.500$); T4 is-a-tumor-visible ($0.500$); and T5 lesion laterality
(left/right/bilateral/none, $0.250$). Each task is posed in a multiple-choice
(MC) format with randomized option order; the model is instructed to give an
answer \emph{and} state a confidence in $0$--$100$; missing or unparseable
answers are counted as non-coverage. Each task is additionally posed open-ended
(OE). For OE responses, we count a hallucinated finding when the model asserts a
tumor, lesion, edema, mass effect, abnormal laterality, or other pathology on an
input whose source label, mask, or control status provides no support for that
finding; refusals, uncertainty statements, and non-pathological descriptions are
not counted as hallucinations.

\noindent\textbf{Models.} We audit six instruction-tuned VLMs with deterministic
(greedy) decoding: the general-purpose
InternVL2.5-8B~\cite{chen2024internvl25} and
Qwen2.5-VL-3B/7B~\cite{bai2025qwen25vl}; Gemma-3-4B, which serves as the
\emph{base} model~\cite{gemmateam2025gemma3}; Gemma-4-12B, the newest model in the
Gemma family that we audit~\cite{gemma4_12b}; and MedGemma-4B, a \emph{medical
fine-tune} of Gemma-3-4B~\cite{sellergren2025medgemma}. Two architectural details
matter for interpretation. MedGemma-4B retains the Gemma-3 language backbone but
its vision encoder (MedSigLIP) was further trained on medical image--text
pairs~\cite{sellergren2025medgemma}; the base-vs-medical contrast therefore provides a family-level probe of medical
adaptation while holding the language backbone fixed, recognizing that the
vision-side training pipeline also differs. Gemma-4-12B,
by contrast, uses an \emph{encoder-free} vision pathway in which image patches are
projected directly into the language backbone rather than through a SigLIP-style
encoder~\cite{gemma4_12b}; we keep this in mind when comparing it to the other
models. A seventh model, InternVL2.5-2B, was degenerate (it produced a parseable
MC answer on essentially no items) and is excluded from all results.

\noindent\textbf{Metrics.} For each (model, task) we report coverage (the fraction
of gradeable items answered); accuracy and \emph{balanced
accuracy}~\cite{brodersen2010balanced}; ECE (10-bin, equal-width)~\cite{naeini2015obtaining} and the Brier
score~\cite{brier1950verification}; the mean confidence on wrong answers and the
\emph{confidently-wrong rate} (the fraction of answered items that are both
incorrect and high-confidence, $\geq$0.8); and, on negative controls, abstention
appropriateness. Confidence intervals come from a bootstrap that resamples whole
subjects (1{,}000 resamples)~\cite{efron1993bootstrap}, because slices and prompts
from the same subject are statistically correlated. Throughout, we audit
\emph{verbalized} confidence---the 0--100 value the model states, not internal
token probabilities or a full predictive distribution---which is the signal a
deployer actually sees; and because a model reports confidence only for its
selected answer, the Brier score is computed on the binary event that the selected
answer is correct. Because raw accuracy pools tasks with different chance levels
and class distributions, we treat it as a descriptive aggregate and use task-level
balanced accuracy as the main competence measure.

\begin{table*}[t]
  \caption{Reliability summary (neutral prompts, MC). ``Conf.\ wrong'' is the mean
  stated confidence on incorrect answers; ``CW-rate'' is the fraction of answered
  items that are high-confidence ($\geq$0.8) \emph{and} wrong. Lower is better
  for ECE, Brier, Conf.\ wrong, and CW-rate. Subject-clustered 95\% bootstrap
  intervals are computed for robustness but omitted from this short-format table.}
  \label{tab:reliability}
  \small
  \begin{tabular}{lrrrrrr}
    \toprule
    Model & Coverage & Accuracy & ECE $\downarrow$ & Brier $\downarrow$ & Conf.\ wrong $\downarrow$ & CW-rate $\downarrow$\\
    \midrule
    InternVL2.5-8B & 1.000 & 0.567 & 0.361 & 0.371 & 0.917 & 0.426\\
    Qwen2.5-VL-3B  & 1.000 & 0.514 & 0.361 & 0.355 & 0.844 & 0.357\\
    Qwen2.5-VL-7B  & 1.000 & 0.563 & \textbf{0.272} & 0.326 & \textbf{0.819} & 0.360\\
    Gemma-3-4B     & 1.000 & 0.531 & 0.404 & 0.405 & 0.922 & 0.462\\
    Gemma-4-12B    & 0.996 & \textbf{0.670} & 0.307 & \textbf{0.311} & 0.968 & \textbf{0.330}\\
    MedGemma-4B    & 1.000 & 0.571 & 0.381 & 0.389 & 0.949 & 0.428\\
    \bottomrule
  \end{tabular}
\end{table*}

\begin{table}[t]
  \caption{Compact task-resolved patterns (neutral prompts, MC). We report
  ranges and representative magnitudes rather than the full model-by-task matrix.
  BA~=~balanced accuracy.}
  \label{tab:bytask}
  \small
  \renewcommand{\arraystretch}{1.16}
  \begin{tabular}{@{}p{0.29\columnwidth} >{\centering\arraybackslash}p{0.12\columnwidth} p{0.45\columnwidth}@{}}
    \toprule
    \textbf{Task pattern} & \textbf{Chance BA} & \textbf{Observed pattern}\\
    \midrule
    Brain-vs-non-brain easiest & 0.50 & General models high (BA $0.91$--$0.99$); MedGemma lower ($0.66$).\\
    \addlinespace[3pt]
    Laterality fails & 0.25 & All models near chance (BA $0.23$--$0.35$).\\
    \addlinespace[3pt]
    Plane is model-dependent & 0.33 & Most near chance/moderate (BA $0.33$--$0.48$); Gemma-4-12B higher ($0.74$).\\
    \addlinespace[3pt]
    Tumor presence is mixed & 0.50 & BA ranges $0.45$--$0.67$; MedGemma improves over its base ($0.58$ vs.\ $0.46$).\\
    \addlinespace[3pt]
    Accuracy $\neq$ safer confidence & -- & The most accurate model has the highest confidence on wrong answers.\\
    \bottomrule
  \end{tabular}
\end{table}

\section{Results}
\textbf{Coverage and calibration.} Five of the six models answer \emph{every}
gradeable MC item (coverage $=1.000$); Gemma-4-12B answers $0.996$ and is the only
model that ever declines or emits an unparseable answer. High coverage is itself
safety-relevant: these models almost never refrain from answering, so their
reliability rests entirely on whether the answers---and the attached
confidences---are trustworthy. Table~\ref{tab:reliability} shows they are not. ECE
ranges from $0.272$ (Qwen2.5-VL-7B) to $0.404$ (Gemma-3-4B), and the mean
confidence \emph{on wrong answers} ranges from $0.819$ to $0.968$. Between roughly
one third and one half of all answered items are high-confidence errors
(confidently-wrong rate $0.330$--$0.462$). In short, incorrect answers are frequently accompanied by high stated
confidence.

\noindent\textbf{Task competence vs.\ chance.} The task-resolved pattern
(Table~\ref{tab:bytask}) is as informative as the aggregate. Brain-vs-non-brain
detection was the only consistently reliable task for the general models (balanced
accuracy above $0.90$); the medical model was the notable exception, far lower, as
it over-accepted inputs---including non-brain controls---as brain MRI. Lesion
laterality stayed close to its $0.25$ chance level for every model: none reliably
localized a lesion to a hemisphere from a single slice. Sequence and tumor-presence
judgments were mixed and varied strongly by model family, and imaging-plane
recognition was solved only by the newest model. Reporting one aggregate accuracy
would hide all of this structure.

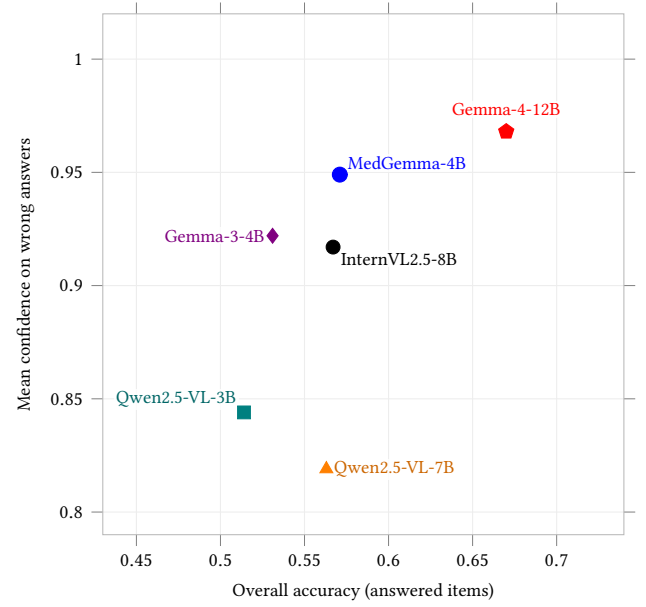
\begin{figure}[t]
\centering
\begin{tikzpicture}
\begin{axis}[
  width=1\columnwidth,
  height=1\columnwidth,
  xlabel={Overall accuracy (answered items)},
  ylabel={Mean confidence on wrong answers},
  xmin=0.43, xmax=0.74, ymin=0.79, ymax=1.02,
  grid=both, grid style={gray!14}, tick align=outside,
  tick label style={font=\footnotesize}, label style={font=\footnotesize},
  axis line style={gray!55}, clip=false,
]
\addplot[only marks, mark=*,         mark size=2.6pt, color=black]  coordinates {(0.567,0.917)};
\addplot[only marks, mark=square*,   mark size=2.4pt, color=teal]   coordinates {(0.514,0.844)};
\addplot[only marks, mark=triangle*, mark size=2.8pt, color=orange] coordinates {(0.563,0.819)};
\addplot[only marks, mark=diamond*,  mark size=2.8pt, color=violet] coordinates {(0.531,0.922)};
\addplot[only marks, mark=pentagon*, mark size=3.0pt, color=red]    coordinates {(0.670,0.968)};
\addplot[only marks, mark=oplus*,    mark size=2.8pt, color=blue]   coordinates {(0.571,0.949)};

\node[anchor=south,      font=\footnotesize, text=red,
      fill=white, fill opacity=0.8, text opacity=1, inner sep=0.8pt, yshift=5pt]
      at (0.670,0.968) {Gemma-4-12B};

\node[anchor=south west, font=\footnotesize, text=blue,
      fill=white, fill opacity=0.8, text opacity=1, inner sep=0.8pt, xshift=2pt, yshift=1pt]
      at (0.571,0.949) {MedGemma-4B};

\node[anchor=north west, font=\footnotesize, text=black,
      fill=white, fill opacity=0.8, text opacity=1, inner sep=0.8pt, xshift=2pt, yshift=-1pt]
      at (0.567,0.917) {InternVL2.5-8B};

\node[anchor=east,       font=\footnotesize, text=violet,
      fill=white, fill opacity=0.8, text opacity=1, inner sep=0.8pt, xshift=-2pt]
      at (0.531,0.922) {Gemma-3-4B};

\node[anchor=south east, font=\footnotesize, text=teal!80!black,
      fill=white, fill opacity=0.8, text opacity=1, inner sep=0.8pt, xshift=-2pt, yshift=1pt]
      at (0.514,0.844) {Qwen2.5-VL-3B};

\node[anchor=west,       font=\footnotesize, text=orange!80!black,
      fill=white, fill opacity=0.8, text opacity=1, inner sep=0.8pt, xshift=2pt]
      at (0.563,0.819) {Qwen2.5-VL-7B};
\end{axis}
\end{tikzpicture}
\caption{Overall multiple-choice accuracy versus mean stated confidence on incorrect answers for the six audited VLMs (neutral prompts). Higher accuracy does not imply safer confidence behavior: Gemma-4-12B is the most accurate model overall, but it also has the highest confidence on its errors. Qwen2.5-VL-7B is less accurate but substantially less overconfident when wrong.}
\label{fig:scatter}
\Description{Scatter plot of overall multiple-choice accuracy against mean confidence on incorrect answers for six vision-language models: InternVL2.5-8B, Qwen2.5-VL-3B, Qwen2.5-VL-7B, Gemma-3-4B, Gemma-4-12B, and MedGemma-4B. Gemma-4-12B is the rightmost and highest point, indicating the highest accuracy and the highest confidence on wrong answers.}
\end{figure}

\noindent\textbf{Competence does not buy calibration.} Gemma-4-12B, the newest
model, is the most accurate overall (accuracy $0.670$ vs.\ $0.514$--$0.571$ for
the rest) and the strongest on sequence, plane, and---narrowly---laterality. Yet
it is simultaneously the \emph{most overconfident on its errors}: its mean
confidence on wrong answers is $0.968$, the highest of any model, and its ECE
($0.307$) is worse than Qwen2.5-VL-7B's ($0.272$) despite far higher accuracy
(Fig.~\ref{fig:scatter}; Table~\ref{tab:reliability}). Its confidently-wrong \emph{rate} is the lowest
($0.330$) only because it errs less often---but the errors it does make are stated
with near-maximal confidence. Because Gemma-4-12B also uses a different
(encoder-free) vision pathway, we read this not as a clean ``scale increases
overconfidence'' law but as a more basic message: \emph{the strongest model we
audited illustrates that higher accuracy need not bring safer confidence
behavior.}
Progress on accuracy should not be read as progress on reliability; the two must
be measured separately.

\noindent\textbf{Family-level effect of medical fine-tuning.} Because MedGemma-4B
is a medical fine-tune of Gemma-3-4B with the same language backbone, their
difference offers a family-level contrast between a base model and its medical
adaptation---recognizing that the adaptation also changes the vision-side training
pipeline, not the language backbone alone. The effect is a \emph{trade}, not a
uniform gain: medical adaptation improved tumor-presence detection but reduced
performance on generic visual tasks such as imaging-plane and brain-vs-non-brain
recognition, and did not improve confidence-on-error. A plausible mechanism is
that retuning the vision encoder on medical data sharpened lesion sensitivity at
the expense of the generic visual discrimination these tasks require---consistent
with reports that medical multimodal LLMs lose visual fidelity relative to their
base models~\cite{zhu2026lost}.

\begin{table}[t]
  \caption{Secondary OE/OOD diagnostics. Hallucination is measured on
  unsupported open-ended inputs; abstention is appropriateness on non-brain
  controls.}
  \label{tab:secondary}
  \small
  \begin{tabular}{lcc}
    \toprule
    Model & OE halluc. $\downarrow$ & Abstention $\uparrow$\\
    \midrule
    InternVL2.5-8B & 0.031 & 0.589\\
    Qwen2.5-VL-3B  & 0.084 & 0.368\\
    Qwen2.5-VL-7B  & 0.097 & 0.629\\
    Gemma-3-4B     & 0.130 & 0.739\\
    Gemma-4-12B    & 0.127 & 0.750\\
    MedGemma-4B    & 0.061 & 0.504\\
    \bottomrule
  \end{tabular}
\end{table}

\noindent\textbf{Secondary diagnostics: hallucination and abstention.} With the
open-ended (OE) pass complete for all six models, hallucinated findings on
unsupported inputs are non-zero across the panel (Table~\ref{tab:secondary}).
The two Gemma models have the highest hallucination rates (Gemma-3-4B $0.130$,
Gemma-4-12B $0.127$), while Gemma-4-12B is also the strongest multiple-choice
model. Abstention on non-brain controls is highly variable rather than uniformly
poor, ranging from $0.368$ for Qwen2.5-VL-3B to $0.750$ for Gemma-4-12B. Format
also shifts measured competence: open-ended querying tends to raise sequence
accuracy but lower tumor-presence accuracy. Together these show that accuracy,
verbalized-confidence reliability, hallucination, and abstention are
\emph{distinct behavioral axes}: the most accurate model is also among the most
overconfident on its errors and the most hallucinatory in open text, yet the most
willing to abstain---so collapsing them into one safety score would mislead. The
calibration failure is not idiosyncratic to our setting: our ECE range is broadly
consistent with that reported for inference-only medical VLMs on unrelated VQA
benchmarks~\cite{khanmohammadi2026calibrated}.

\section{Discussion and Implications}
The six-model audit supports a coherent and sobering picture. \emph{Confidence is
decoupled from correctness:} coverage is near-complete and stated confidence is
high, but accuracy is modest and calibration is poor across the board, with mean
confidence on wrong answers of $0.82$--$0.97$. For an unsupervised assistant, this is the dangerous regime: well-formed, confident wrong answers with little self-doubt. \emph{Capability gains
did not transfer to calibration:} the most accurate model is the most
overconfident on its mistakes, and among the most prone to open-ended
hallucination. \emph{Medical fine-tuning is a targeted trade, not
a free lunch:} in this family-level comparison, medical adaptation improved tumor
detection but degraded generic recognition and did not improve overconfidence---a
medical label does not by itself imply broad visual competence or safe confidence
behavior. And \emph{task structure matters:} most general models reliably distinguish brain from non-brain, the strongest
model shows task-dependent gains in sequence and plane recognition, tumor
presence remains mixed, and single-slice laterality remains near chance. Together these
motivate the paper's central recommendation: medical-image VLM evaluation should
report coverage, calibration, confident error, and abstention \emph{alongside}
accuracy, because a model can improve on accuracy while remaining---or
becoming---less safe. Concretely, this argues for safety-audit reporting beyond accuracy-only leaderboards: coverage, calibration, confident-error analysis, and abstention under distribution shift. For AI leadership, procurement, and oversight, the
implication is direct: independent calibration and abstention audits should accompany accuracy claims
before a vision--language model is considered for clinical-triage workflows.

\noindent\textbf{Limitations.} The headline uses a single 2-D slice and an MC answer; richer context might raise accuracy, but more context alone does not address calibration or abstention. We audit \emph{self-reported}
confidence as a deployer would see it rather than recalibrating it; cross-model
calibration rankings should therefore be read with the caveat that they partly
reflect how each model was trained to express certainty, and Gemma-4-12B's
distinct encoder-free vision pathway is a confound for cross-model comparison.
Tumor labels reflect post-treatment glioma masks and a fixed visibility threshold rather than radiologist re-reads; laterality uses the displayed slice coordinate frame and is an exploratory stress test, not a clinical localization benchmark. The
negative-control set is small (70 images), so abstention estimates are descriptive
rather than definitive. BraTS and IXI are public datasets, so training-data
exposure cannot be ruled out; nevertheless, the audit grades slice-level visual
behavior under controlled prompts rather than subject-level memorization. The
model panel contains six reported models, one medical specialist, and one
base/specialist family pair. Results pertain to this distribution; we make no
clinical-diagnostic claim.

\section{Conclusion}
We presented an automatically graded audit framework requiring no new annotation and pilot evidence on calibration
and confident error for six general and medical VLMs on brain MRI. Models answer
almost everything with high confidence but are poorly calibrated and frequently
confidently wrong; the newest and most accurate model is the most overconfident on
its errors; and a family-level comparison indicates that medical adaptation can trade
generic visual competence for a targeted tumor-detection gain without improving
calibration. The practical implication is that accuracy alone is an inadequate---
and potentially misleading---summary of a medical-image VLM's readiness, and that
calibration, confident error, and abstention must be first-class evaluation
targets.

\section*{Ethics and Privacy Statement}
This study is a behavioral audit conducted entirely on publicly available,
de-identified brain-MRI-derived images (BraTS-2024 and IXI) and synthetic or
non-medical control images; it involves no human subjects, no private data, and no
patient interaction, and all labels are derived programmatically from public
metadata and released segmentation masks. The intended benefit is safer evaluation of medical-imaging AI by exposing confident error and inappropriate non-abstention. The principal risk is misinterpretation---our findings concern a specific
public distribution and must not be read as evidence that any audited model is fit
(or unfit) for clinical use; we make no diagnostic claim and encourage
radiologist-verified evaluation before any deployment decision.

\begin{acks}
Research was sponsored by the Army Research Laboratory and was accomplished under
Cooperative Agreement Number W911NF-23-2-0224. The views and conclusions contained
in this document are those of the authors and should not be interpreted as
representing the official policies, either expressed or implied, of the Army
Research Laboratory or the U.S. Government. The U.S. Government is authorized to
reproduce and distribute reprints for Government purposes notwithstanding any
copyright notation herein.
\end{acks}

\bibliographystyle{ACM-Reference-Format}
\bibliography{refs}

@inproceedings{guo2017calibration,
  author    = {Chuan Guo and Geoff Pleiss and Yu Sun and Kilian Q. Weinberger},
  title     = {On Calibration of Modern Neural Networks},
  booktitle = {Proceedings of the 34th International Conference on Machine Learning},
  series    = {Proceedings of Machine Learning Research},
  volume    = {70},
  pages     = {1321--1330},
  year      = {2017},
  publisher = {PMLR},
  url       = {https://proceedings.mlr.press/v70/guo17a.html}
}

@inproceedings{naeini2015obtaining,
  author    = {Mahdi Pakdaman Naeini and Gregory F. Cooper and Milos Hauskrecht},
  title     = {Obtaining Well Calibrated Probabilities Using Bayesian Binning},
  booktitle = {Proceedings of the Twenty-Ninth AAAI Conference on Artificial Intelligence},
  pages     = {2901--2907},
  year      = {2015},
  publisher = {AAAI Press}
}

@article{brier1950verification,
  author  = {Glenn W. Brier},
  title   = {Verification of Forecasts Expressed in Terms of Probability},
  journal = {Monthly Weather Review},
  volume  = {78},
  number  = {1},
  pages   = {1--3},
  year    = {1950},
  doi     = {10.1175/1520-0493(1950)078<0001:VOFEIT>2.0.CO;2}
}

@inproceedings{brodersen2010balanced,
  author    = {Kay H. Brodersen and Cheng Soon Ong and Klaas E. Stephan and Joachim M. Buhmann},
  title     = {The Balanced Accuracy and Its Posterior Distribution},
  booktitle = {Proceedings of the 20th International Conference on Pattern Recognition},
  pages     = {3121--3124},
  year      = {2010},
  publisher = {IEEE},
  doi       = {10.1109/ICPR.2010.764}
}

@book{efron1993bootstrap,
  author    = {Bradley Efron and Robert J. Tibshirani},
  title     = {An Introduction to the Bootstrap},
  publisher = {Chapman \& Hall/CRC},
  address   = {New York, NY},
  year      = {1993}
}

@article{menze2015brats,
  author  = {Bjoern H. Menze and Andras Jakab and Stefan Bauer and Jayashree Kalpathy-Cramer and Keyvan Farahani and others},
  title   = {The Multimodal Brain Tumor Image Segmentation Benchmark ({BRATS})},
  journal = {IEEE Transactions on Medical Imaging},
  volume  = {34},
  number  = {10},
  pages   = {1993--2024},
  year    = {2015},
  doi     = {10.1109/TMI.2014.2377694}
}

@article{bakas2017advancing,
  author  = {Spyridon Bakas and Hamed Akbari and Aristeidis Sotiras and Michel Bilello and Martin Rozycki and others},
  title   = {Advancing The Cancer Genome Atlas Glioma {MRI} Collections with Expert Segmentation Labels and Radiomic Features},
  journal = {Scientific Data},
  volume  = {4},
  pages   = {170117},
  year    = {2017},
  doi     = {10.1038/sdata.2017.117}
}

@misc{deverdier2024brats,
  author        = {Maria Correia de Verdier and Rachit Saluja and Louis Gagnon and Dominic LaBella and Ujjwal Baid and others},
  title         = {The 2024 Brain Tumor Segmentation ({BraTS}) Challenge: Glioma Segmentation on Post-treatment {MRI}},
  year          = {2024},
  archiveprefix = {arXiv},
  eprint        = {2405.18368},
  primaryclass  = {cs.CV},
  doi           = {10.48550/arXiv.2405.18368}
}

@misc{ixi,
  author       = {{Biomedical Image Analysis Group, Imperial College London}},
  title        = {{IXI} Dataset: Information eXtraction from Images},
  year         = {2026},
  howpublished = {\url{https://brain-development.org/ixi-dataset/}},
  note         = {Dataset page, accessed June 29, 2026}
}

@misc{bai2025qwen25vl,
  author        = {Shuai Bai and Keqin Chen and Xuejing Liu and Jialin Wang and Wenbin Ge and others},
  title         = {{Qwen2.5-VL} Technical Report},
  year          = {2025},
  archiveprefix = {arXiv},
  eprint        = {2502.13923},
  primaryclass  = {cs.CV},
  doi           = {10.48550/arXiv.2502.13923}
}

@misc{chen2024internvl25,
  author        = {Zhe Chen and Weiyun Wang and Yue Cao and Yangzhou Liu and Zhangwei Gao and others},
  title         = {Expanding Performance Boundaries of Open-Source Multimodal Models with Model, Data, and Test-Time Scaling},
  year          = {2024},
  archiveprefix = {arXiv},
  eprint        = {2412.05271},
  primaryclass  = {cs.CV},
  doi           = {10.48550/arXiv.2412.05271}
}

@misc{gemmateam2025gemma3,
  author        = {{Gemma Team}},
  title         = {{Gemma 3} Technical Report},
  year          = {2025},
  archiveprefix = {arXiv},
  eprint        = {2503.19786},
  primaryclass  = {cs.CL},
  doi           = {10.48550/arXiv.2503.19786},
  note          = {Google DeepMind}
}

@misc{sellergren2025medgemma,
  author        = {Andrew Sellergren and Sahar Kazemzadeh and Tiam Jaroensri and Atilla Kiraly and others},
  title         = {{MedGemma} Technical Report},
  year          = {2025},
  archiveprefix = {arXiv},
  eprint        = {2507.05201},
  primaryclass  = {cs.AI},
  doi           = {10.48550/arXiv.2507.05201},
  note          = {Google Research and Google DeepMind}
}

@misc{gemma4_12b,
  author       = {Olivier Lacombe and Gus Martins},
  title        = {Introducing {Gemma 4} 12{B}: A Unified, Encoder-Free Multimodal Model},
  year         = {2026},
  howpublished = {Google Blog},
  url          = {https://blog.google/innovation-and-ai/technology/developers-tools/introducing-gemma-4-12b/},
  note         = {Released June 3, 2026; accessed June 29, 2026}
}

@inproceedings{li2023pope,
  author    = {Yifan Li and Yifan Du and Kun Zhou and Jinpeng Wang and Wayne Xin Zhao and Ji-Rong Wen},
  title     = {Evaluating Object Hallucination in Large Vision-Language Models},
  booktitle = {Proceedings of the 2023 Conference on Empirical Methods in Natural Language Processing},
  pages     = {292--305},
  year      = {2023},
  publisher = {Association for Computational Linguistics},
  address   = {Singapore},
  doi       = {10.18653/v1/2023.emnlp-main.20},
  url       = {https://aclanthology.org/2023.emnlp-main.20/}
}

@misc{byun2026overconfidence,
  author        = {Ji Young Byun and Young-Jin Park and Jean-Philippe Corbeil and Asma Ben Abacha},
  title         = {Overconfidence and Calibration in Medical {VQA}: Empirical Findings and Hallucination-Aware Mitigation},
  year          = {2026},
  archiveprefix = {arXiv},
  eprint        = {2604.02543},
  primaryclass  = {cs.CV},
  doi           = {10.48550/arXiv.2604.02543}
}

@misc{khanmohammadi2026calibrated,
  author        = {Reza Khanmohammadi and Kundan Thind and Mohammad M. Ghassemi},
  title         = {Calibrated Triage, Not Autonomy: Confidence Estimation for Medical Vision-Language Models},
  year          = {2026},
  archiveprefix = {arXiv},
  eprint        = {2606.15910},
  primaryclass  = {cs.CL},
  doi           = {10.48550/arXiv.2606.15910}
}

@misc{zhu2026lost,
  author        = {Xun Zhu and Fanbin Mo and Xi Chen and Kaili Zheng and Shaoshuai Yang and Yiming Shi and Jian Gao and Miao Li and Ji Wu},
  title         = {Lost in the Hype: Revealing and Dissecting the Performance Degradation of Medical Multimodal Large Language Models in Image Classification},
  year          = {2026},
  archiveprefix = {arXiv},
  eprint        = {2604.08333},
  primaryclass  = {cs.CV},
  doi           = {10.48550/arXiv.2604.08333}
}

\end{document}